\documentclass[sigconf,nonacm]{acmart}

\usepackage{graphicx}
\usepackage[ruled,noend,noline,linesnumbered]{algorithm2e}
\usepackage{multicol}
\usepackage{multirow}
\usepackage{amsmath}
\usepackage{multicol}
\usepackage{mathtools}
\usepackage{booktabs}

\DeclarePairedDelimiter{\floor}{\lfloor}{\rfloor}
\DeclareMathOperator*{\argmax}{arg\,max}

\newcommand\Alpha{\mathrm{A}}
\graphicspath{ {./img/} }

\begin{document}

\title{Competing in a Complex Hidden Role Game with Information Set Monte Carlo Tree Search}

\author{Jack Reinhardt}
\email{reinh240@umn.edu}
\affiliation{%
  \institution{University of Minnesota}
  \city{Minneapolis}
  \state{Minnesota}
  \postcode{55455}
}

%

\begin{abstract}
Advances in intelligent game playing agents have led to successes in perfect information games like \textit{Go} and imperfect information games like \textit{Poker}. The Information Set Monte Carlo Tree Search (ISMCTS) family of algorithms outperforms previous algorithms using Monte Carlo methods in imperfect information games. In this paper, Single Observer Information Set Monte Carlo Tree Search (SO-ISMCTS) is applied to \textit{Secret Hitler}, a popular social deduction board game that combines traditional hidden role mechanics with the randomness of a card deck. This combination leads to a more complex information model than the hidden role and card deck mechanics alone. It is shown in 10108 simulated games that SO-ISMCTS plays as well as simpler rule based agents, and demonstrates the potential of ISMCTS algorithms in complicated information set domains.
\end{abstract}


\maketitle

%

\section{Introduction}\label{sec:intro}

\if@twocolumn
  Hello World!
\fi

The use of board games in artificial intelligence research is a popular and powerful tool in developing new algorithms \cite{russell_artificial_2009}. Board games have a set of predefined rules to govern the play of the game, and therefore provide a simpler domain for algorithm development than complex real-world domains. Although many algorithms were originally developed in the context of board games, they have been applied to real-world problems from oil transportation \cite{trunda_using_2013} to interplanetary travel \cite{hennes_interplanetary_2015}.\par 

Past research in artificial intelligence has resulted in algorithms with the ability to beat the best human players in games such as \textit{Chess}, and recently with DeepMind's AlphaGo algorithm, \textit{Go} \cite{silver_mastering_2017}. \textit{Chess} and \textit{Go}, however, provide an algorithm with all of the information in the game. In real-world contexts, perfect information like this is rare. Therefore, it is important to research algorithms that do not require perfect information. Many board games feature imperfect information: \textit{Blackjack} and \textit{Poker} for example have a hidden card deck. Previous research has created algorithms that are able to play games with hidden information \cite{cowling_information_2012, cowling_emergent_2015,eger_keeping_2018, ginsberg_gib_2001, serrino_finding_2019}.\par

Other board games hide information in different ways, one such mechanic is known as hidden roles. In hidden role games, the identity of certain players is hidden from other players. Examples of these games include \textit{Werewolf}, \textit{Mafia}, and \textit{The Resistance}. Artificial intelligence algorithms have also been created to play these types of games \cite{eger_keeping_2018,serrino_finding_2019}.\par

This paper explorers the effectiveness of existing imperfect information algorithm Single Observer Information Set Monte Carlo Tree Search (SO-ISMCTS) on the popular board game \textit{Secret Hitler}. Secret Hitler is a social deduction game that combines traditionally separate imperfect information domains -- hidden role and card deck mechanics. The combination of hidden roles and randomness of a card deck gives the game a more complex information model than previously studied games like \textit{The Resistance} or \textit{Werewolf}. The added complexity provides a challenging test for available imperfect information agents: opponents' moves might be driven by ulterior motives stemming from their hidden role, or simply forced by the random card deck.\par

There are many existing algorithms for imperfect information domains. Examples include Counterfactual Regret Minimization \cite{neller_introduction_2013}, Epistemic Belief Logic \cite{eger_keeping_2018}, and  Information Set Monte Carlo Tree Search \cite{cowling_information_2012, cowling_emergent_2015}. Due to its ability to handle large state spaces and produce results in a variable amount of computation time \cite{cowling_information_2012}, Single Observer Information Set Monte Carlo Tree Search (SO-ISMCTS) is chosen as the main algorithm to apply to Secret Hitler.\par

In section \ref{sec:secrethitler}, an overview of the game mechanics and rules are presented. In section \ref{sec:relatedwork} recent work on similar games is explored. In section \ref{sec:agents}, the algorithms (also referred to as agents) to play the game are described. Section \ref{sec:performance} details experimental results of these algorithms. In section \ref{sec:futurework}, future work in this domain is noted.\par

%

\section{The Game}\label{sec:secrethitler}

\textit{Secret Hitler} is a board game of political intrigue and betrayal, set in 1930s Germany during the Pre-World War II Weimar Republic. The game was created in 2016 after a successful Kickstarter campaign raising over 1.4 million USD, and has since risen in popularity \cite{bromwich_secret_2017,max_temkin_secret_2018}. It is the 6\textsuperscript{th} ranked hidden role game on \href{https://boardgamegeek.com/boardgamemechanic/2891/hidden-roles}{boardgamegeek.com} as of April 2020 \cite{boardgamegeek_hidden_role}. It introduces a new policy deck mechanic to the hidden role genre, leading to other interesting mechanics with positions of power explained in section \ref{subsec:rules}. The genre has gained popularity due to games like \textit{Mafia}, and \textit{The Resistance} \cite{maranges_hidden_2015}.\par

The theme of Secret Hitler may be provocative for some readers due to the atrocities that transpired during Adolf Hitler and the Nazi Party's Third Reich. The mechanics of the game are not affected by the theme, and could apply to any nation featuring a partisan government: it could very well be named \textit{Secret President}. Secret Hitler takes place during the time which Adolf Hitler rose to power, and explorers the dynamics within the government. It does not ``model the specifics of German parliamentary politics'' \cite{max_temkin_secret_2018} or simulate historically accurate policies enacted by the government. On their Kickstarter page, the game-makers state that Secret Hitler ``model[s] the paranoia and distrust [Hitler] exploited, the opportunism that his rivals failed to account for, and the disastrous temptation to solve systemic problems by giving more power to the `right people'\thinspace'' \cite{max_temkin_secret_2018}. Secret Hitler's new game mechanics provide a new challenge for artificial intelligence agents worth exploring, despite its unconventional theme.\par

\begin{figure}[htbp]
    \centering
    \includegraphics[width=\columnwidth]{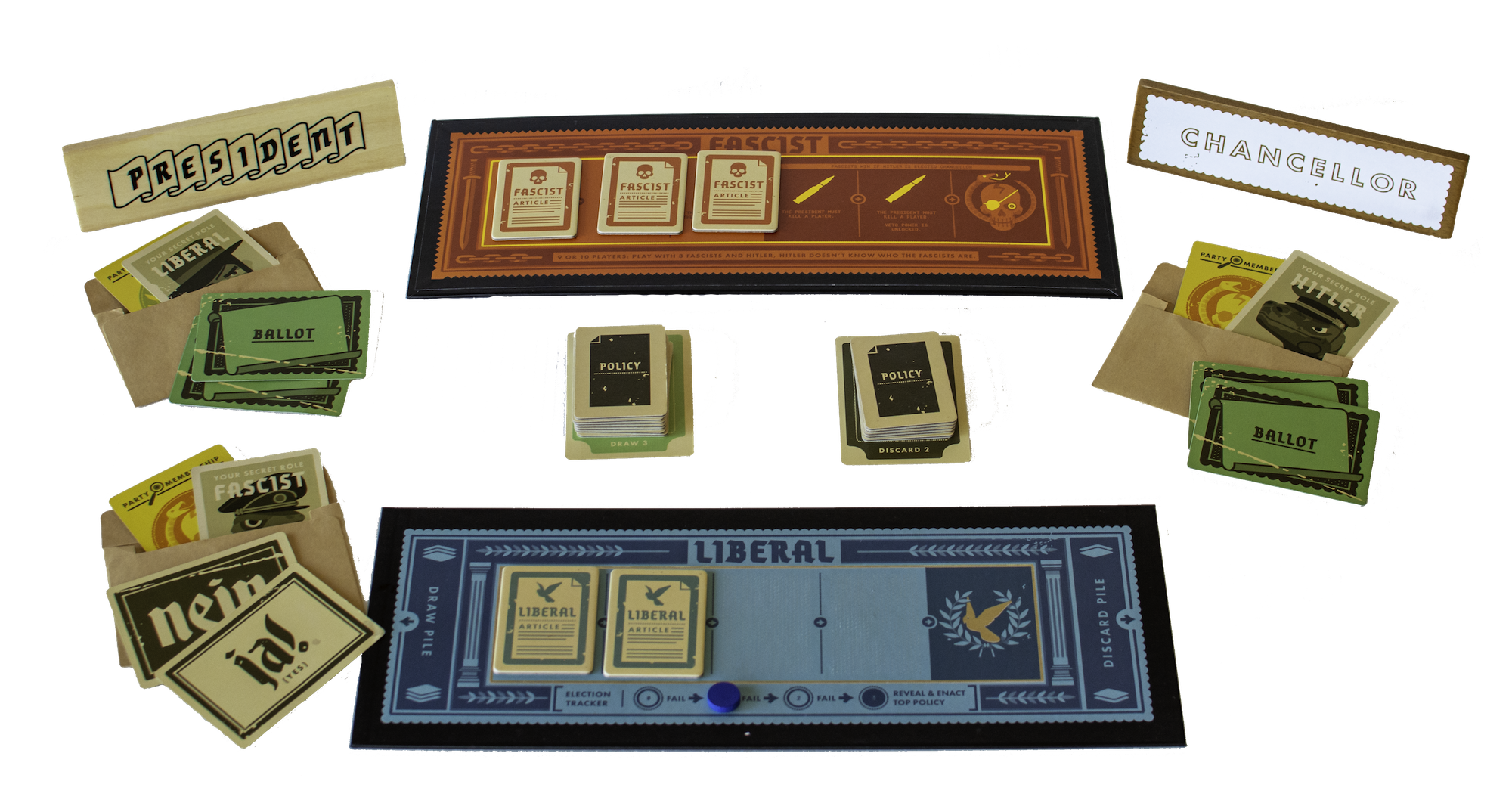}
    \caption{The Secret Hitler board mid-game. There are three Fascist policies enacted, and two Liberal policies enacted. The chaos marker is positioned at 1. Secret role cards are not shown during the game, but shown here for demonstration. Ballot cards, however, are visible throughout the game. The current President and Chancellor markers are also shown in the upper portion of the figure.}
    \label{fig:secrethitlergame}
\end{figure}

\subsection{Rules and Procedures}\label{subsec:rules}
Players of Secret Hitler are thrust into the Reichstag, the legislative house of the Weimar Republic, and assume one of three secret roles: Liberal, Fascist, or Hitler. The Fascists know the identity of their teammates and Hitler. The Liberals and Hitler do not know the identity of anyone.\par

The game is played with 5-10 players; the number of Fascists and Liberals varies depending on the number of players. The Liberal team always has a narrow majority. Table \ref{table:numfasplayers} maps the number of total players to the number of Fascist players.\par

\begin{table}[htbp]
\caption{Number of Fascist Players Given Total Players}
\begin{tabular}{ c c }\toprule
 Total Players & Fascists \\ \hline
 5 - 6 & 1 + Hitler \\
 7 - 8 & 2 + Hitler \\
 9 - 10 & 3 + Hitler \\\bottomrule
\end{tabular}
\label{table:numfasplayers}
\end{table}

\subsubsection{Objective}

The goal of the game is to enact Liberal or Fascist policy cards to either a) save the country from fascism, or b) overthrow the government and create a fascist regime. Policy cards only specify the party of the policy, not any actual government policy. They serve as a score-keeping device throughout the game. There are four ways to win:

\makeatletter%
\if@twocolumn\else
	\begin{multicols}{2}
\fi
\begin{itemize}
    \item Enact 6 Fascist policies. \textbf{Fascists win.}
    \item Hitler is elected Chancellor. \textbf{Fascists win.}
    \item Enact 5 Liberal policies. \textbf{Liberals win.}
    \item Hitler is killed. \textbf{Liberals win.}
\end{itemize}
\makeatletter%
\if@twocolumn\else
	\end{multicols}
\fi

\subsubsection{Game Play}
The game is broken up into multiple legislative sessions (rounds). In each session, a player is assigned the President title. The current President must nominate one of the other players as Chancellor. The President-Chancellor pair is referred to as a \textit{Proposed Government}.\par

To legitimatize the Proposed Government, all players must vote either \textit{Ja} (yes), or \textit{Nein} (no). These votes are simultaneous and publicly known. If the Proposed Government passes with a simple majority, the Government is legitimatized. If the Proposed Government fails, the legislative session closes and the Presidency is rotated to the next player at the table.\par

\begin{figure*}[ht]
\makeatletter%
\if@twocolumn%
	\includegraphics[width=0.9\textwidth]{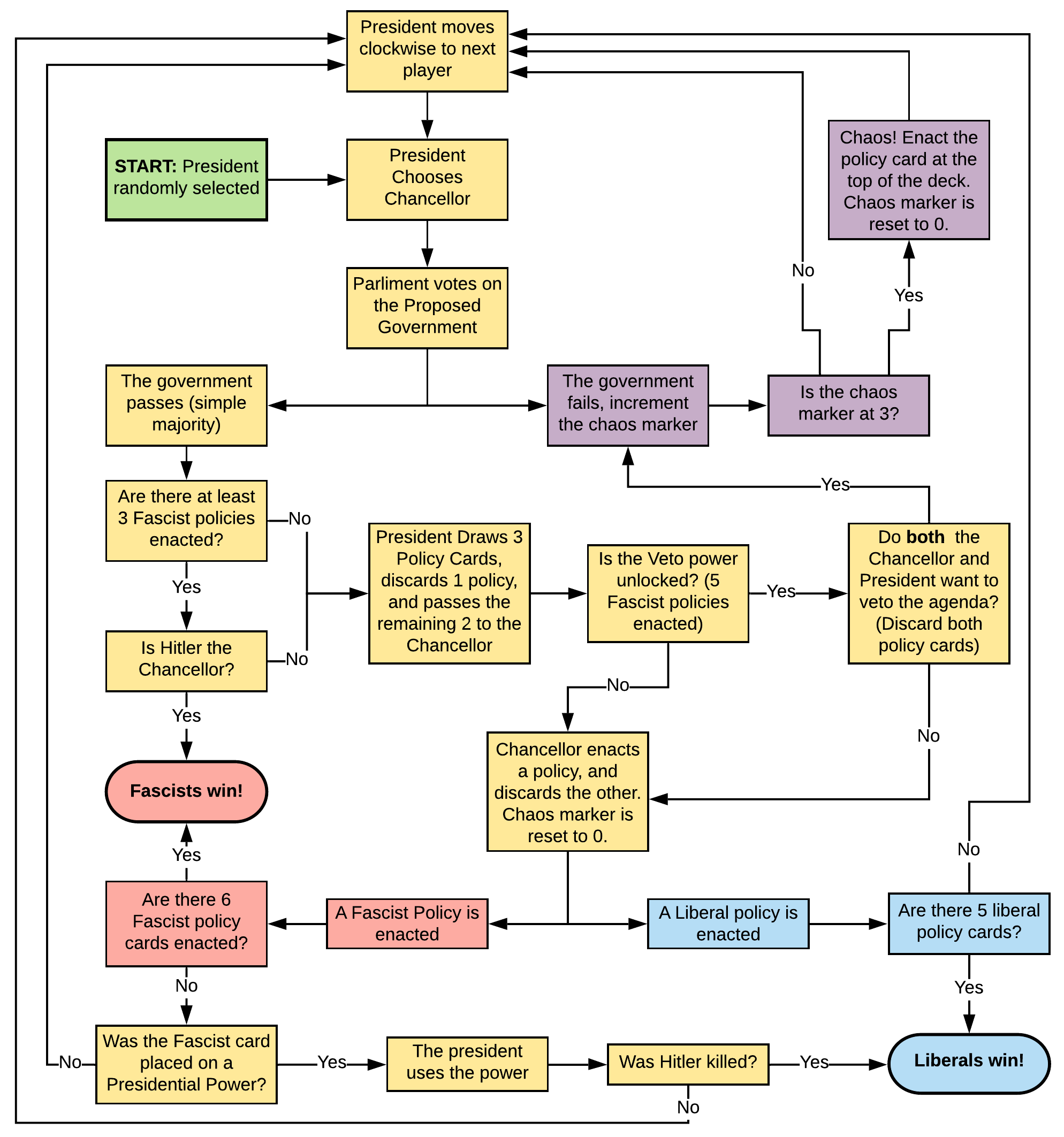}
\else
	\includegraphics[width=\textwidth]{sh_gameplay_flowchart}
\fi
\caption{Flow of Play Throughout Secret Hitler}
\label{fig:flowofsh}
\end{figure*}

If the Government is passed, the President draws three policy cards from the deck, and discards one. The President passes the two remaining policy cards to the Chancellor. The Chancellor then chooses which policy card to play (the other policy card is discarded). Only the enacted policy is visible to the table, the other cards are discarded secretly. The Presidency then rotates to the next player at the table. Refer to Figure \ref{fig:flowofsh} for a complete description of the flow of the game.\par

\subsubsection{Presidential Powers}

Some Fascist policy spaces on the board contain a Presidential Power. The Power is granted to the sitting President once a Fascist policy card is placed on the corresponding space. The powers include: Investigate Loyalty, Call Special Election, Policy Peek, Execution, and Veto. Refer to table \ref{table:prespowers} in Appendix \ref{app:prespowers} for a description of the powers.\par

\subsubsection{Other Details}

Some details in the description of Secret Hitler were omitted for brevity and clarity; important caveats are listed here. 
\begin{itemize}
\item The Fascist team will win if Hitler is elected Chancellor \textbf{and} three or more Fascist policies have been enacted.
\item Hitler knows the identity of the other fascist in 5 and 6 player games.
\item If 3 consecutive Proposed Governments fail, chaos ensues. The top policy card is immediately played on the board.
\end{itemize}

%

\section{Related Work}\label{sec:relatedwork}

In past years, intelligent agents have had success in games like \textit{Chess} and \textit{Go} \cite{silver_mastering_2017}. However, in these games, the agent does not have to determine who the opponent is. Social Deduction games (also referred to as hidden role games) provide this extra challenge. Recently, agents have been developed for the hidden role games \textit{One Night Ultimate Werewolf} and \textit{The Resistance: Avalon} \cite{eger_keeping_2018, serrino_finding_2019}.\par

Eger and Martens have developed an agent for \textit{One Night Ultimate Werewolf} that utilizes Dynamic Epistemic Logic to model the beliefs of the agent. They experiment with different levels of commitment to the current belief and find that the optimal commitment level depends on the opponent's strategy \cite{eger_keeping_2018}. Dynamic Epistemic Logic to infer beliefs is not explored in this paper.\par

Serrino et al. have developed DeepRole, an agent for \textit{The Resistance: Avalon}, that utilizes a variant of Counterfactual Regret Minimization and deep value networks. They were able to test their agent against other agents in over 20,000 simulated games, in addition to over 2000 games against human players on \href{https://www.proavalon.com/}{ProAvalon.com}. They found that DeepRole had a higher win rate than other agents in simulated games, in addition to outperforming human players \cite{serrino_finding_2019}. \textit{The Resistance: Avalon} and Secret Hitler have similar game mechanics, with Secret Hitler adding the randomness of a policy deck. The DeepRole algorithm may be able to effectively play Secret Hitler, however it is not implemented here.\par

Variants on the popular Monte Carlo Tree Search (MCTS) algorithm have been developed for games with imperfect information, such as Bridge \cite{ginsberg_gib_2001}. Perfect Information Monte Carlo (PIMC) sampling is a simple variant of MCTS that can handle imperfect information. This algorithm uses determinization to create many perfect information game trees. It then averages the resulting payoff of each move at the root node across trees to choose an action. Long et al. have analyzed the performance of PIMC and found that though it has been criticized for its theoretical deficiencies, in practice it can lead to strong agents \cite{long_understanding_2010}.\par

Cowling et al. developed another variant of MCTS - Information Set Monte Carlo Tree Search (ISMCTS) - and explore its use in several domains of imperfect information \cite{cowling_information_2012,cowling_emergent_2015}. They found that their algorithms Multiple Observer ISMCTS, Multiple Tree ISMCTS, and Single Observer ISMCTS outperform previous algorithms utilizing Monte Carlo methods. The latter is adapted for use in Secret Hitler in the following sections.\par

It is traditionally thought that communication in cooperative games is imperative to success \cite{eger_intentional_2017}. The challenge in communicating while playing hidden role games arises from the imperfect information held by each player. The player must determine who to cooperate with, and who they must oppose. Only then can they effectively communicate with their supposed teammates. Players of hidden role or social deduction games will notice the high level of verbal communication between human players. Few agents developed so far have been able to communicate with teammates or adversaries directly, but the ability to do so provides an advantage to an agent \cite{eger_intentional_2017}. Communication with other agents or human players is not considered in this paper.\par

%

\section{Agent Design}\label{sec:agents}

Three agents were created to play Secret Hitler: Random, Selfish, and Single Observer Monte Carlo Tree Search (SO-ISMCTS). These agents fall into two categories: rule based agents, and Monte Carlo Tree Search (MCTS) agents. The rule based agents were developed to compete with the more intelligent and general purpose MCTS agent. They also serve as a performance benchmark for future algorithms to play Secret Hitler.\par

There are many algorithms that have been created for games featuring imperfect information \cite{cowling_information_2012,cowling_emergent_2015,eger_keeping_2018,serrino_finding_2019}. This paper chooses to investigate the effectiveness of the MCTS variant Single Observer Information Set Monte Carlo Tree Search (SO-ISMCTS) in the context of the Secret Hitler game. MCTS is effective in games with large branching factors and a large state space. Secret Hitler has a tree size of at least $10^{60}$ and a hidden state size of $10^5$. See Appendix \ref{app:statespace} these calculations.\par

It was originally thought that the existing online implementation of the game at \href{https://secrethitler.io}{secrethitler.io} could be used within the algorithms developed here to model the game rules. This idea was eventually discarded due to efficiency and practicality concerns. Instead, a native python implementation of the game was developed for the following algorithms.\par

\subsection{Rule Based Agents}

Rule based agents (also referred to as \textit{knowledge-intensive} agents) are simple, and can be quite effective. However, they are highly specialized and require extensive domain knowledge to implement. They become ineffective once outside the domain they were designed for \cite{russell_artificial_2009}.\par

Although domain specific rule based agents are not particularly elegant solutions, they are useful in providing benchmarks for more intelligent agents. Here, a two rule based agents are developed, namely a Random Agent and a Selfish Agent.\par

\subsubsection{Random}

The random agent is self explanatory. It randomly chooses an action from the set of legal actions, even if the action would cause the agent to lose the game. This is the simplest agent, and all other agents created should aim to outperform this agent.

\subsubsection{Selfish}

The Selfish agent is identical to the random agent, except in cases where legal actions involve enacting or discarding a policy. The agent will always attempt to enact its own party's policy, and discard the opposing party's policy. For example, if the agent is a Liberal President, it will discard a fascist policy (if possible). See Algorithm \ref{alg:selfish} for pseudo-code.\par

This agent should perform better than the random agent because it will not intentionally choose a policy that results in a loss. However, given enough policies enacted, the party affiliation of this agent can easily be determined by an adversary.\par

\begin{algorithm}[ht]
\DontPrintSemicolon
\SetKwProg{function}{function}{}{}
\SetKwFunction{selfish}{Selfish}

\function{\selfish{$\Alpha$, $\pi$}}{
    \If{$EnactPolicy(\pi) \in \Alpha$}{
    \KwRet $EnactPolicy(\pi)$\;
    }
    \If{$DiscardPolicy(\neg\pi) \in \Alpha$}{
    \KwRet $DiscardPolicy(\neg\pi)$\;
    }
    choose $\alpha \in \Alpha$ uniformly at random\;
    \KwRet $\alpha$\;
}
\caption[]{The Selfish Algorithm. The following notation is used:
\begin{itemize}
    \item $\Alpha =$ set of legal actions
    \item $\pi =$ party affiliation of the agent. $\neg\pi$ represents the opposing party to the agent
\end{itemize}
}
\label{alg:selfish}
\end{algorithm}

\subsection{Information Set Monte Carlo Tree Search}

In past years, MCTS algorithms have been successful in solving games with perfect information like Go and Chess \cite{silver_mastering_2017}. Unlike more traditional AI heuristic algorithms, MCTS does not require an evaluation function. MCTS instead simulates many random games from the current game state, assigning payoff values to each state it encounters based on the result of the random game. These payoff values are averaged throughout many simulations, and therefore an informed action can be made at the current state \cite{gelly_monte-carlo_2011}.\par

Information Set Monte Carlo Tree Search (ISMCTS) is a family of MCTS algorithms developed for imperfect information domains \cite{cowling_information_2012,cowling_emergent_2015}. Instead of searching a tree of game states, these algorithms search a tree of information sets. Information sets are a collection of states which cannot be distinguished by a player due to hidden information. For example in a card game, an information set might contain game states with all permutations of the opponent's cards. The player knows which information set they are in, but not the game state within the information set. In this paper, Single Observer Information Set Monte Carlo Tree Search (SO-ISMCTS) is applied to Secret Hitler.\par

\subsection{Single Observer Information Set Monte Carlo Tree Search (SO-ISMCTS)}

Single Observer Information Set Monte Carlo Tree Search (SO-ISMCTS) is part of the Information Set Monte Carlo Tree Search (ISMCTS) family of algorithms. It therefore searches a tree of information sets instead of game states. Nodes in the tree are information sets, and edges in the tree represent actions. This variation of ISMCTS gets its name because it maintains one tree from one player's point of view.\par

At a high level, SO-ISMCTS works by picking a random determinization of the hidden state for $n$ iterations, and running MCTS on the information set tree. Concretely, a determinization in a card game would be \textit{one} permutation of the opponent's cards. SO-ISMCTS avoids some of the deficiencies of Perfect Information Monte Carlo Sampling due to the sharing of nodes between determinizations instead of creating a new tree for each determinization. Here, we use $n = 10000$ determinizations. Algorithm \ref{alg:soismcts} in Appendix \ref{app:algorithms} details the pseudo-code of SO-ISMCTS from \cite{cowling_information_2012}.\par

Some modifications to SO-ISMCTS were required to apply it to Secret Hitler due to the randomization of the policy deck. As the algorithm is selecting and expanding nodes in the tree, it is advancing through different game states and hidden states. At some point, the policy deck in the hidden state is depleted, and the policies in the discard pile are shuffled to form the new policy deck. This process is not deterministic, and causes problems with the \textsc{Backpropagate} function. During backpropagation, there is no guarantee the deck will be shuffled in the same way as during selection and expansion. Because of this, some moves that were valid in selection and expansion are no longer valid in backpropagation.\par

To remedy this problem, a queue of game states and hidden states is maintained in each determinization of the algorithm. Instead of using the game rules to transition to a new game state in backpropagation, the new states are fetched from the queue. This enables deterministic transitions between hidden states. In this way, each action in the action history of the current node will be valid with the given game state and hidden state. In other words, it allows for backpropagation to assign rewards to the correct nodes. The solution does not modify the nodes themselves, thus preserving the node sharing properties of SO-ISMCTS.\par

A small optimization was made to SO-ISMCTS to shrink the number of possible hidden states. As the game progresses, some determinizations become more or less likely. At later stages in the game, some hidden states become impossible due to observations in the game. For example, in Secret Hitler, if player $x$ is successfully elected chancellor while at least 3 fascist policies are enacted, the observing player can eliminate hidden states in which player $x$ is Hitler. SO-ISMCTS takes advantage of this fact and maintains a list of possible hidden role configurations. At every move, it filters out impossible hidden role configurations from its internal list.\par

An important consideration in SO-ISMCTS is how to implement random determinization of hidden states. In Secret Hitler, there are around $10^5$ possible hidden states (See Appendix \ref{app:statespace} for details). At the start of every iteration of the algorithm, a determinization must be generated. Since this operation is done many times, it is beneficial to make this operation as fast as possible.\par

A naive approach might be to generate all hidden states, and randomly choose from this list. This approach is computationally expensive and can be optimized by using lazy evaluation. Each component of the hidden state is eagerly evaluated, but the combination of these components is lazily evaluated. Concretely, all of the possible hidden roles, and policy decks are computed (these are pre-processed at the start of the game). The components are randomized at the start of SO-ISMCTS, and every time the agent requests a new determinization, one is created on the fly.\par

%

\section{Performance}\label{sec:performance}

To test the performance of the agents, each agent played in simulated games against the other agents. Overall win rate is the primary metric that this paper will use to judge the effectiveness of each agent. The win rates for each agent as the number of players change, and win rates for different assignments of secret roles are explored. These win rates offer a more granular evaluation of the performance of each agent than the overall win rate; they provide potential explanations for the overall performance of each agent.\par

\subsection{Configuration}\label{subsec:performanceconfig}

Since Secret Hitler is not a two-player game, comparing each agent head-to-head is difficult. The win rate of each agent could vary depending on the number of players in the game, the agent's teammates, and the agent's opponents. Additionally, there are three possible secret roles an agent can be assigned. The win rate might vary based on the agent's assigned secret role, or even the secret roles of the other agents. The configurations for each variable is shown in Table \ref{table:config}.\par

\begin{table}[htbp]
    \centering
    \caption{Secret Hitler Configurations}
    \begin{tabular}{l l}\toprule
         Parameter & Possible Values \\ \hline
         Number of Players & 5 - 10 \\
         Agents & Random, Selfish, SO-ISMCTS \\
         Secret Role & Liberal, Fascist, Hitler \\ \bottomrule
    \end{tabular}
    \label{table:config}
\end{table}

There are 6 variants of the game, one for each of the possible number of players from 5 to 10. There are a total of 3 agents: Random, Selfish, and SO-ISMCTS. If $n$ is the number of players, the 3 agents can be assigned in $3+n-1 \choose n$ ways. Hitler can be assigned $n$ ways. The remaining fascists $f = \floor{\frac{n-1}{2}} - 1$ can be assigned $n-1 \choose f$ ways. So, for each variant of the game, we have that the number of configurations $C = \sum_{n=5}^{10} {3+n-1 \choose n} \times n \times {n-1 \choose f} \approx 10^4$. It is, therefore, not feasible to explicitly define all configurations.\par

To solve this problem, the configuration in each game is randomized; this is identical to choosing a configuration uniformly at random for each game. Over a large number of games played, the expected amount times each configuration is chosen is equal. Therefore, an overall win-rate metric is suitable in determining the performance of each agent.\par

\subsection{Example Game}

To concretely demonstrate the testing process for these agents, an example game is presented. In this game, the configurations were randomly assigned as in Table \ref{table:example}.\par

\begin{table}[htbp]
    \centering
    \caption{Example Game Configuration. Their last name initial represents their Secret Role.}
    \begin{tabular}{l l l}\toprule
         Player Name & Algorithm & Role     \\ \hline
         Grace F.   & Random    & Fascist   \\ 
         Eve L.     & Random    & Liberal   \\ 
         Alice H.   & SO-ISMCTS & Hitler    \\ 
         Bob F.     & SO-ISMCTS & Fascist   \\ 
         Michael L. & Selfish   & Liberal   \\ 
         Victor L.  & Selfish   & Liberal   \\ 
         Carol L.   & SO-ISMCTS & Liberal   \\ 
         Frank L.   & SO-ISMCTS & Liberal   \\ \bottomrule 
    \end{tabular}
    \label{table:example}
\end{table}

The game starts with Frank L. as President, who chooses Alice H. as their Chancellor. The last name initial of these characters represents their secret role. The ensuing election fails as only Grace F., Carol L. and Frank L. vote \textit{Ja}. The chaos marker is incremented from 0 to 1 and the presidency moves to Grace F. The next 2 elections also fail, and the top policy from the deck is enacted; the policy is Liberal.\par

The next election is successful with Alice H. as president and Bob F. as chancellor. Alice H. draws two Fascist policies and one Liberal policy from the deck, and discards one of the Fascist policies. Bob F. then enacts the Fascist policy. The score is now tied at 1.\par

Alice H.'s choice to discard a Fascist policy goes against their party affiliation. Many human players would grade this move poorly; they would have discarded the Liberal policy. They could claim that they handed the chancellor two Fascist policies because they drew 3 Fascist policies from the deck. Some players might say that since Alice H. does not know who the other Fascist players are, they could be "testing" Bob F. to glean information about their secret role with this move. However, it is not very beneficial to analyze the intent behind each specific move the agent makes as humans assume an agent is intentional \cite{dennett_intentional_1971}.\par

Play continues until the next successful election in which Alice H. is again President, and Grace F. is Chancellor. Alice H. draws three policies, passes two to Grace F., who enacts a Fascist policy. This gives the President the power to investigate a player's party membership. Alice H., who does not know the identity of the other Fascists, chooses to investigate Frank L. Alice H. finds that Frank L. is a Liberal, and in the future will only explore determinizations where Frank L. is a Liberal.\par

The next interesting move occurs later in the game, when the fourth Fascist policy is enacted; Carol L. is President and Grace F. is Chancellor. The fourth Fascist policy in this game configuration gives the president the Execution power - they must kill a player (remove them from the game). Carol L. chooses to execute Eve L. Eve L. can no longer participate in the game. Since Eve L. is not Hitler, the game continues, and all players know that Eve L. is not Hitler.\par

The next election fails, but in the following election, Alice H. is successfully elected as Chancellor, and the game ends in a Fascist victory. This result is not uncommon. See Table \ref{table:win_reasons} for a breakdown of how games typically end. There are many more interesting ways the game can progress, as there are at least $10^{60}$ possible games, but only one game is analyzed in detail here. See Appendix \ref{app:statespace} for tree size calculation.\par

\begin{table}[htbp]
    \centering
    \caption{Frequency of Game Ending Reasons.}
    \begin{tabular}{l c}\toprule
         Game Ending Reason     & Probability   \\ \hline
         Hitler Elected         & \textbf{0.477}\\
         Six Fascist Policies   & 0.251         \\
         Five Liberal Policies  & 0.154         \\
         Hitler Killed          & 0.118         \\ \bottomrule
    \end{tabular}
    \label{table:win_reasons}
\end{table}

\subsection{Win Rate Analysis}

\begin{figure}[htbp]
    \centering
    \if@twocolumn%
        \includegraphics[width=\columnwidth]{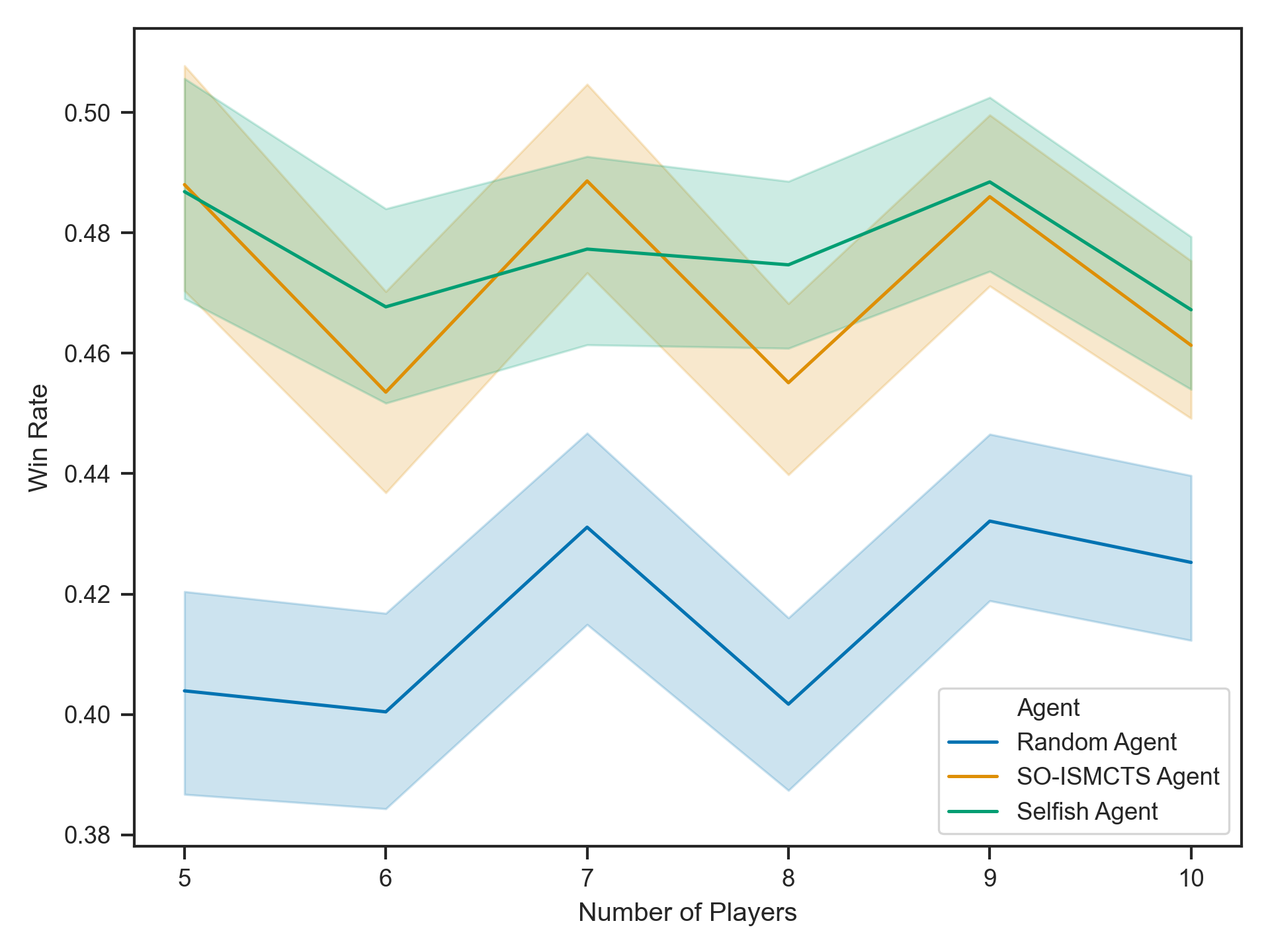}
    \else
        \includegraphics[width=0.75\columnwidth]{img/by_number_players_win_rate.png}
    \fi
    \caption{Agent Win Rate by Number of Players. Shaded regions represent 95\% Confidence Intervals. Lowest line is the Random agent, the flattest line is the Selfish Agent, and the most erratic line is the SO-ISMCTS agent.}
    \label{fig:player_win_rate}
\end{figure}

10108 games were run using random configurations described in \ref{subsec:performanceconfig}. Table \ref{table:win_rates} contains the overall win rate for each agent throughout all games. Table \ref{table:win_rates_secret_role} contains win rates and  for each agent, and their secret role. Figure \ref{fig:player_win_rate} shows the win rate for each agent as the number of players changes A 95\% confidence interval is calculated for all win rate metrics.

Table \ref{table:win_rates} shows that the Selfish agent has the highest overall win rate at $47.7\%$ of games won. Its true win rate is higher than the Random agent's true win rate with 95\% confidence. However, its win rate is not significantly higher than the SO-ISMCTS agent, as the confidence intervals overlap. Since the Selfish and Random agents are simple rule-based agents, they make decisions extremely quickly. SO-ISMCTS on the other hand takes up to 20 seconds to choose an action. The best overall algorithm for an agent competing in the games conducted here is the Selfish agent due to its decision making speed, and that it is not significantly worse than the SO-ISMCTS agent.\par

\begin{table}[htbp]
    \centering
    \caption{Overall Win Rates for each agent. Confidence interval notation: $(lower\;bound,\;upper\;bound)$.}
    \begin{tabular}{ c c c }\toprule
         Agent      & Win Rate & 95\% CI     \\ \hline
         SO-ISMCTS  & 0.471  & (0.465, 0.477) \\
         Selfish    & \textbf{0.477}  & \textbf{(0.471, 0.483)} \\
         Random     & 0.417  & (0.411, 0.423) \\ \bottomrule
    \end{tabular}
    \label{table:win_rates}
\end{table}

The games simulated here do not have any human opponents. The Selfish agent would likely do worse against human opponents, because its party affiliation (if Fascist) is easily exposed. No agent developed here analyzes the past moves of its opponents to determine the party affiliation of its opponents, so it makes sense why the selfish agent has success here.\par

Table \ref{table:win_rates_secret_role} shows that the SO-ISMCTS agent has the highest win rate of all agents when playing as a Liberal or Fascist, but not as Hitler. Interestingly, SO-ISMCTS performs worse than the Random agent when playing as Hitler; this is likely due to SO-ISMCTS being reluctant to enact Fascist policies as Hitler. Since some of its opponents here act randomly, it may happen that a Fascist policy placed on the Execution Presidential Power will lead to Hitler's execution (potentially by a teammate), which results in a loss. A SO-ISMCTS agent playing as Hitler may prefer to enact a Liberal policy in these cases, giving the Liberal team a greater likelihood to win. However, this hypothesis is not confirmed, and more testing needs to be done to determine why this relative dip in performance exists.\par

\begin{table*}[htbp]
    \centering
    \caption{Win Rates for each agent by Secret Role. Confidence interval notation: $(lower\;bound,\;upper\;bound)$.}
    \begin{tabular}{ c c c c c c c }\toprule
        \multirow{2}{*}{Agent} & \multicolumn{6}{c}{Win Rate} \\ \cmidrule{2-7}
        & Liberal & 95\% CI & Fascist & 95\% CI & Hitler & 95\% CI \\ \hline
        SO-ISMCTS  & \textbf{0.287} & \textbf{(0.280, 0.295)} & \textbf{0.773} & \textbf{(0.763, 0.783)} & 0.697 & (0.682, 0.713) \\
        Selfish    & 0.286 & (0.279, 0.293)  & 0.754  & (0.744, 0.765) & \textbf{0.776}  & \textbf{(0.762, 0.790)} \\
        Random     & 0.231 & (0.224, 0.237)  & 0.695  & (0.684, 0.706) & 0.712  & (0.696, 0.727) \\ \bottomrule
    \end{tabular}
    \label{table:win_rates_secret_role}
\end{table*}

Figure \ref{fig:player_win_rate} depicts the win rate of each agent against the number of players in the game. The Random agent is significantly worse than the Selfish and SO-ISMCTS agents in all variants of the game. The Selfish agent and SO-ISMCTS agents, though close in overall win rate, have different behaviors as the number of players changes. The Selfish agent seems to have a more consistent win rate as the number of players varies, whereas the SO-ISMCTS and Random agents' win rates are more erratic.\par

This interesting behavior is likely due to the ratio of Liberal players and Fascist players in each variant of the game. With an odd number of players (an odd game), the ratio of Liberal to Fascist players is small: the Liberal team has \textbf{one} more player than the Fascist team. With an even number of players (an even game), the ratio of Liberal to Fascist players increases: the Liberal team has \textbf{two} more players than the Fascist team.\par

The erratic win rate of the SO-ISMCTS and Random agents could be due to the low overall win rate of Liberal players seen in Table \ref{table:win_rates_secret_role}. It is more likely that an agent playing in an even game is a Liberal, and hence will have a lower win rate than in odd games. However, this does not account for the relative stability of the Selfish agent's win rate across odd and even games. More testing is necessary here to uncover the reason the Selfish agent's win rate is more consistent than the other agent's win rates.\par

\subsection{High-Performance Cloud Computing}

To enable mass quantities of games to be played in a short period of time, cloud computing using Google Kubernetes Engine was utilized. A $12$ preemptible node cluster was created with $89$ vCPUs, allowing for high parallelization at a low cost. Kubernetes Jobs representing each game were run in parallel, with the aforementioned randomized configurations. Upon completion of each game, the results were written to a MongoDB instance also running on Kubernetes. This enabled efficient querying of the win rates of each agent.\par

%

\section{Future Work}\label{sec:futurework}

It has been demonstrated here that the Single Observer Information Set Monte Carlo Tree Search (SO-ISMCTS) algorithm fares no better than the Selfish agent in games with Random, Selfish, and SO-ISMCTS opponents. SO-ISMCTS is unable to predict the secret role of agents, and is limited in its bluffing ability when compared to agents developed for other games. Implementing other variants of ISMCTS, like Multiple Observer ISMCTS or Many Tree ISMCTS \cite{cowling_information_2012, cowling_emergent_2015} could lead to a higher win rate. Other types of algorithms could also be applied to Secret Hitler, like Counterfactual Regret Minimization \cite{neller_introduction_2013}, or a variant of DeepRole \cite{serrino_finding_2019}.\par

Games against humans are valuable in determining the strength of an agent against a wide variety of strategies. Humans are unpredictable and may cause agents that play well against other computer programs to perform worse than expected. Future work should test agents against human opponents. Appendix \ref{app:online} references a popular online version of Secret Hitler in which an interface for agents to play could be developed.\par

Future research could also investigate why the Selfish agent's win rate is less erratic than the SO-ISMCTS and Random agents when the number of players is varied. It is an intriguing pattern that likely stems from the variable number of Liberals and Fascists in a specific game configuration.\par 

The algorithms produced here do not communicate directly with one another. In physical games, and the online game, communication is thought to be vital. Future agents could explore the benefit of participating in the chat throughout the game. This could lead to more explicit alliances throughout the game and may enable a higher level of bluffing.\par

%

\section{Conclusions}\label{sec:conclusion}

In this paper, Random, Selfish, and Single Observer Information Set Monte Carlo Tree Search (SO-ISMCTS) agents were applied to the popular board game \textit{Secret Hitler}. Secret Hitler introduces new mechanics to the hidden role board game genre that provide a challenging test for existing algorithms that are built to handle imperfect information. The policy deck mechanic injects randomness into the play of each agent. It is more difficult to determine the secret role of a player because their actions could be intentional, or forced by the randomness of the policy deck.\par

SO-ISMCTS and the Selfish agents outperformed the Random agent in a large amount of simulated games. SO-ISMCTS and the Selfish agents did not have a significant difference in overall win rate, which was likely due to the small amount of opponent strategies in the simulated games. If more strategies were introduced, or if the agents played against human opponents, these results will likely change. Interestingly, the win rates of each agent were not constant across secret role assignments and the number of opponents. Fascist players tend to have higher win rates than liberal players, and SO-ISMCTS and Random agents have erratic changes in win rates with different numbers of opponents.\par

Although SO-ISMCTS did not have a significantly different win rate than the Selfish agent, it has shown the potential of Information Set Monte Carlo Tree Search (ISMCTS) algorithms to outperform rule based agents in Secret Hitler. Some of the shortcomings of SO-ISMCTS have been resolved with Multiple Observer ISMCTS and Many Tree ISMCTS, so it is likely these agents will yield a better win rate in Secret Hitler than SO-ISMCTS.\par

In the future, research into hidden role games provides an exciting opportunity for artificial intelligence agents. Games in this genre simulate the complex social interactions that humans encounter on a daily basis. Questions like who to trust, how to call other's bluffs, and what to communicate to other members of a group are all questions present in hidden role games. Developing effective algorithms for hidden role games allows us to explore human social interactions to a degree not seen in other games.\par

%

\begin{acks}
The author would like to thank Maria Gini for all of the helpful conversations and guidance during the development of this work. They would like to thank Shana Watters and Kuen-Bang Hou (Favonia) for feedback during the writing process. They would also like to thank Chris Ozols and the \href{https://secrethitler.io}{Secret Hitler.io} development community for initial help in playing the online game programmatically, though not used in this paper.
\end{acks}

\bibliographystyle{ACM-Reference-Format}
\bibliography{references}


\begin{thebibliography}{18}


\ifx \showCODEN    \undefined \def \showCODEN     #1{\unskip}     \fi
\ifx \showDOI      \undefined \def \showDOI       #1{#1}\fi
\ifx \showISBNx    \undefined \def \showISBNx     #1{\unskip}     \fi
\ifx \showISBNxiii \undefined \def \showISBNxiii  #1{\unskip}     \fi
\ifx \showISSN     \undefined \def \showISSN      #1{\unskip}     \fi
\ifx \showLCCN     \undefined \def \showLCCN      #1{\unskip}     \fi
\ifx \shownote     \undefined \def \shownote      #1{#1}          \fi
\ifx \showarticletitle \undefined \def \showarticletitle #1{#1}   \fi
\ifx \showURL      \undefined \def \showURL       {\relax}        \fi
\providecommand\bibfield[2]{#2}
\providecommand\bibinfo[2]{#2}
\providecommand\natexlab[1]{#1}
\providecommand\showeprint[2][]{arXiv:#2}

\bibitem[\protect\citeauthoryear{BoardGameGeek}{BGG}{2020}]%
        {boardgamegeek_hidden_role}
BGG \bibinfo{year}{2020}\natexlab{}.
\newblock \bibinfo{title}{Hidden Roles {\textbar} Board Game Mechanic
  {\textbar} BoardGameGeek}.
\newblock
\newblock
\urldef\tempurl%
\url{https://boardgamegeek.com/boardgamemechanic/2891/hidden-roles}
\showURL{%
Retrieved April 28, 2020 from \tempurl}


\bibitem[\protect\citeauthoryear{Bromwich}{Bromwich}{2017}]%
        {bromwich_secret_2017}
\bibfield{author}{\bibinfo{person}{Jonah~Engel Bromwich}.}
  \bibinfo{year}{2017}\natexlab{}.
\newblock \bibinfo{title}{Secret Hitler, a Game That Simulates Fascism’s
  Rise, Becomes a Hit}.
\newblock
\newblock
\urldef\tempurl%
\url{https://www.nytimes.com/2017/09/05/business/secret-hitler-game.html}
\showURL{%
Retrieved Februrary 19, 2020 from \tempurl}


\bibitem[\protect\citeauthoryear{Cowling, Whitehouse, and Powley}{Cowling
  et~al\mbox{.}}{2012}]%
        {cowling_information_2012}
\bibfield{author}{\bibinfo{person}{Peter~I. Cowling}, \bibinfo{person}{Daniel
  Whitehouse}, {and} \bibinfo{person}{Edward~J. Powley}.}
  \bibinfo{year}{2012}\natexlab{}.
\newblock \showarticletitle{Information Set Monte Carlo Tree Search}.
\newblock \bibinfo{journal}{\emph{IEEE Transactions on Computational
  Intelligence and AI in Games}} \bibinfo{volume}{4}, \bibinfo{number}{2}
  (\bibinfo{date}{June} \bibinfo{year}{2012}), \bibinfo{pages}{120--143}.
\newblock
\showISSN{1943-068X, 1943-0698}
\urldef\tempurl%
\url{https://doi.org/10.1109/TCIAIG.2012.2200894}
\showDOI{\tempurl}


\bibitem[\protect\citeauthoryear{Cowling, Whitehouse, and Powley}{Cowling
  et~al\mbox{.}}{2015}]%
        {cowling_emergent_2015}
\bibfield{author}{\bibinfo{person}{Peter~I. Cowling}, \bibinfo{person}{Daniel
  Whitehouse}, {and} \bibinfo{person}{Edward~J. Powley}.}
  \bibinfo{year}{2015}\natexlab{}.
\newblock \showarticletitle{Emergent bluffing and inference with Monte Carlo
  Tree Search}. In \bibinfo{booktitle}{\emph{2015 IEEE Conference on
  Computational Intelligence and Games (CIG)}}. \bibinfo{publisher}{IEEE},
  \bibinfo{address}{Tainan, Taiwan}, \bibinfo{pages}{114--121}.
\newblock
\showISBNx{978-1-4799-8622-4}
\urldef\tempurl%
\url{https://doi.org/10.1109/CIG.2015.7317927}
\showDOI{\tempurl}


\bibitem[\protect\citeauthoryear{Dennett}{Dennett}{1971}]%
        {dennett_intentional_1971}
\bibfield{author}{\bibinfo{person}{D.~C. Dennett}.}
  \bibinfo{year}{1971}\natexlab{}.
\newblock \showarticletitle{Intentional Systems}.
\newblock \bibinfo{journal}{\emph{Journal of Philosophy}} \bibinfo{volume}{68},
  \bibinfo{number}{4} (\bibinfo{date}{Feb.} \bibinfo{year}{1971}),
  \bibinfo{pages}{87--106}.
\newblock
\showISSN{0022362X}
\urldef\tempurl%
\url{https://doi.org/10.2307/2025382}
\showDOI{\tempurl}


\bibitem[\protect\citeauthoryear{Eger and Martens}{Eger and Martens}{2018}]%
        {eger_keeping_2018}
\bibfield{author}{\bibinfo{person}{Markus Eger} {and} \bibinfo{person}{Chris
  Martens}.} \bibinfo{year}{2018}\natexlab{}.
\newblock \showarticletitle{Keeping the Story Straight: A Comparison of
  Commitment Strategies for a Social Deduction Game}. In
  \bibinfo{booktitle}{\emph{Fourteenth Artificial Intelligence and Interactive
  Digital Entertainment Conference}}. \bibinfo{publisher}{AAAI},
  \bibinfo{address}{Edmonton, Alberta}, \bibinfo{pages}{24--30}.
\newblock
\urldef\tempurl%
\url{https://www.aaai.org/ocs/index.php/AIIDE/AIIDE18/paper/view/18095}
\showURL{%
\tempurl}


\bibitem[\protect\citeauthoryear{Eger, Martens, and Córdoba}{Eger
  et~al\mbox{.}}{2017}]%
        {eger_intentional_2017}
\bibfield{author}{\bibinfo{person}{Markus Eger}, \bibinfo{person}{Chris
  Martens}, {and} \bibinfo{person}{Marcela~Alfaro Córdoba}.}
  \bibinfo{year}{2017}\natexlab{}.
\newblock \showarticletitle{An intentional AI for Hanabi}. In
  \bibinfo{booktitle}{\emph{2017 IEEE Conference on Computational Intelligence
  and Games (CIG)}}. \bibinfo{publisher}{IEEE}, \bibinfo{address}{New York, New
  York}, \bibinfo{pages}{68--75}.
\newblock
\urldef\tempurl%
\url{https://doi.org/10.1109/CIG.2017.8080417}
\showDOI{\tempurl}


\bibitem[\protect\citeauthoryear{Gelly and Silver}{Gelly and Silver}{2011}]%
        {gelly_monte-carlo_2011}
\bibfield{author}{\bibinfo{person}{Sylvain Gelly} {and} \bibinfo{person}{David
  Silver}.} \bibinfo{year}{2011}\natexlab{}.
\newblock \showarticletitle{Monte-Carlo tree search and rapid action value
  estimation in computer Go}.
\newblock \bibinfo{journal}{\emph{Artificial Intelligence}}
  \bibinfo{volume}{175}, \bibinfo{number}{11} (\bibinfo{date}{July}
  \bibinfo{year}{2011}), \bibinfo{pages}{1856--1875}.
\newblock
\showISSN{00043702}
\urldef\tempurl%
\url{https://doi.org/10.1016/j.artint.2011.03.007}
\showDOI{\tempurl}


\bibitem[\protect\citeauthoryear{Ginsberg}{Ginsberg}{2001}]%
        {ginsberg_gib_2001}
\bibfield{author}{\bibinfo{person}{M.~L. Ginsberg}.}
  \bibinfo{year}{2001}\natexlab{}.
\newblock \showarticletitle{GIB: Imperfect Information in a Computationally
  Challenging Game}.
\newblock \bibinfo{journal}{\emph{Journal of Artificial Intelligence Research}}
   \bibinfo{volume}{14} (\bibinfo{date}{June} \bibinfo{year}{2001}),
  \bibinfo{pages}{303--358}.
\newblock
\showISSN{1076-9757}
\urldef\tempurl%
\url{https://doi.org/10.1613/jair.820}
\showDOI{\tempurl}


\bibitem[\protect\citeauthoryear{Hennes and Izzo}{Hennes and Izzo}{2015}]%
        {hennes_interplanetary_2015}
\bibfield{author}{\bibinfo{person}{Daniel Hennes} {and} \bibinfo{person}{Dario
  Izzo}.} \bibinfo{year}{2015}\natexlab{}.
\newblock \showarticletitle{Interplanetary Trajectory Planning with Monte Carlo
  Tree Search}. In \bibinfo{booktitle}{\emph{Proceedings of the 24th
  International Conference on Artificial Intelligence}}
  \emph{(\bibinfo{series}{IJCAI’15})}. \bibinfo{publisher}{AAAI Press},
  \bibinfo{address}{Buenos Aires, Argentina}, \bibinfo{pages}{769–775}.
\newblock
\showISBNx{9781577357384}


\bibitem[\protect\citeauthoryear{Long, Sturtevant, Buro, and Furtak}{Long
  et~al\mbox{.}}{2010}]%
        {long_understanding_2010}
\bibfield{author}{\bibinfo{person}{Jeffrey~Richard Long},
  \bibinfo{person}{Nathan~R. Sturtevant}, \bibinfo{person}{Michael Buro}, {and}
  \bibinfo{person}{Timothy Furtak}.} \bibinfo{year}{2010}\natexlab{}.
\newblock \showarticletitle{Understanding the Success of Perfect Information
  Monte Carlo Sampling in Game Tree Search}. In
  \bibinfo{booktitle}{\emph{Twenty-Fourth AAAI Conference on Artificial
  Intelligence}}. \bibinfo{publisher}{AAAI}, \bibinfo{address}{Atlanta,
  Georgia}, \bibinfo{pages}{134--140}.
\newblock
\urldef\tempurl%
\url{https://www.aaai.org/ocs/index.php/AAAI/AAAI10/paper/view/1876}
\showURL{%
\tempurl}


\bibitem[\protect\citeauthoryear{Maranges}{Maranges}{2015}]%
        {maranges_hidden_2015}
\bibfield{author}{\bibinfo{person}{Tommy Maranges}.}
  \bibinfo{year}{2015}\natexlab{}.
\newblock \bibinfo{title}{Hidden Information in Secret Hitler}.
\newblock
\newblock
\urldef\tempurl%
\url{https://medium.com/@tommygents/hidden-information-in-secret-hitler-f71d0251ee82}
\showURL{%
Retrieved February 19, 2020 from \tempurl}


\bibitem[\protect\citeauthoryear{Neller and Lanctot}{Neller and
  Lanctot}{2013}]%
        {neller_introduction_2013}
\bibfield{author}{\bibinfo{person}{Todd~W Neller} {and} \bibinfo{person}{Marc
  Lanctot}.} \bibinfo{year}{2013}\natexlab{}.
\newblock \bibinfo{title}{An Introduction to Counterfactual Regret
  Minimization}.
\newblock
\newblock
\urldef\tempurl%
\url{http://modelai.gettysburg.edu/2013/cfr/cfr.pdf}
\showURL{%
Retrieved January 21, 2020 from \tempurl}


\bibitem[\protect\citeauthoryear{Russell and Norvig}{Russell and
  Norvig}{2009}]%
        {russell_artificial_2009}
\bibfield{author}{\bibinfo{person}{Stuart Russell} {and} \bibinfo{person}{Peter
  Norvig}.} \bibinfo{year}{2009}\natexlab{}.
\newblock \bibinfo{booktitle}{\emph{Artificial Intelligence: A Modern Approach}
  (\bibinfo{edition}{3rd} ed.)}.
\newblock \bibinfo{publisher}{Prentice Hall Press}, \bibinfo{address}{USA}.
\newblock
\showISBNx{0136042597}


\bibitem[\protect\citeauthoryear{Serrino, Kleiman-Weiner, Parkes, and
  Tenenbaum}{Serrino et~al\mbox{.}}{2019}]%
        {serrino_finding_2019}
\bibfield{author}{\bibinfo{person}{Jack Serrino}, \bibinfo{person}{Max
  Kleiman-Weiner}, \bibinfo{person}{David~C. Parkes}, {and}
  \bibinfo{person}{Joshua~B. Tenenbaum}.} \bibinfo{year}{2019}\natexlab{}.
\newblock \bibinfo{title}{Finding Friend and Foe in Multi-Agent Games}.
\newblock
\newblock
\showeprint{1906.02330}
\urldef\tempurl%
\url{https://arxiv.org/abs/1906.02330}
\showURL{%
Retrieved December 26, 2019 from \tempurl}


\bibitem[\protect\citeauthoryear{Silver, Schrittwieser, Simonyan, Antonoglou,
  Huang, Guez, Hubert, Baker, Lai, Bolton, Chen, Lillicrap, Hui, Sifre, van~den
  Driessche, Graepel, and Hassabis}{Silver et~al\mbox{.}}{2017}]%
        {silver_mastering_2017}
\bibfield{author}{\bibinfo{person}{David Silver}, \bibinfo{person}{Julian
  Schrittwieser}, \bibinfo{person}{Karen Simonyan}, \bibinfo{person}{Ioannis
  Antonoglou}, \bibinfo{person}{Aja Huang}, \bibinfo{person}{Arthur Guez},
  \bibinfo{person}{Thomas Hubert}, \bibinfo{person}{Lucas Baker},
  \bibinfo{person}{Matthew Lai}, \bibinfo{person}{Adrian Bolton},
  \bibinfo{person}{Yutian Chen}, \bibinfo{person}{Timothy Lillicrap},
  \bibinfo{person}{Fan Hui}, \bibinfo{person}{Laurent Sifre},
  \bibinfo{person}{George van~den Driessche}, \bibinfo{person}{Thore Graepel},
  {and} \bibinfo{person}{Demis Hassabis}.} \bibinfo{year}{2017}\natexlab{}.
\newblock \showarticletitle{Mastering the game of Go without human knowledge}.
\newblock \bibinfo{journal}{\emph{Nature}}  \bibinfo{volume}{550}
  (\bibinfo{date}{Oct.} \bibinfo{year}{2017}), \bibinfo{pages}{354--359}.
\newblock
\showISSN{1476-4687}
\urldef\tempurl%
\url{https://doi.org/10.1038/nature24270}
\showDOI{\tempurl}


\bibitem[\protect\citeauthoryear{Temkin}{Temkin}{2018}]%
        {max_temkin_secret_2018}
\bibfield{author}{\bibinfo{person}{Max Temkin}.}
  \bibinfo{year}{2018}\natexlab{}.
\newblock \bibinfo{title}{Secret Hitler}.
\newblock
\newblock
\urldef\tempurl%
\url{https://www.kickstarter.com/projects/maxtemkin/secret-hitler}
\showURL{%
Retrieved April 6, 2020 from \tempurl}


\bibitem[\protect\citeauthoryear{Trunda and Barták}{Trunda and
  Barták}{2013}]%
        {trunda_using_2013}
\bibfield{author}{\bibinfo{person}{Otakar Trunda} {and} \bibinfo{person}{Roman
  Barták}.} \bibinfo{year}{2013}\natexlab{}.
\newblock \showarticletitle{Using Monte Carlo Tree Search to Solve Planning
  Problems in Transportation Domains}. In \bibinfo{booktitle}{\emph{Advances in
  Soft Computing and Its Applications}} \emph{(\bibinfo{series}{Lecture Notes
  in Computer Science})}. \bibinfo{publisher}{Springer},
  \bibinfo{address}{Berlin, Heidelberg}, \bibinfo{pages}{435--449}.
\newblock
\showISBNx{978-3-642-45111-9}
\urldef\tempurl%
\url{https://doi.org/10.1007/978-3-642-45111-9_38}
\showDOI{\tempurl}


\end{thebibliography}

%
\newpage
\appendix

\section{Online Implementation}\label{app:online}

A open source web based implementation of Secret Hitler is available at \href{https://secrethitler.io}{secrethitler.io}. There are slight modifications to the rules to better suit the online nature of the game. These include additional presidential powers, and a slight modification to the veto power. Future agents could connect to this site to evaluate their performance against human opponents.\par

\section{Presidential Powers}\label{app:prespowers}

\begin{table}[ht]
\caption{Presidential Powers and Their Function}
\begin{tabular}{ c p{0.5\columnwidth} }\toprule
 Presidential Power & Description \\ \hline
 Investigate Loyalty    & The President investigates a player's identity card \\
 Call Special Election  & The President picks the next presidential candidate \\
 Policy Peek            & The President examines the top three cards \\
 Execution              & The President must kill a player \\ \bottomrule
\end{tabular}
\label{table:prespowers}
\end{table}

\section{State Space}\label{app:statespace}
The size of the game tree depends on the number of players in the game. Here, only a calculation of the 5 player version is calculated. We can construct a lower bound on the tree size by just considering the longest possible game with 5 players \cite{serrino_finding_2019}.\par

The longest possible game would result in 10 policies enacted: 4 Liberal and 6 Fascist or 5 Liberal and 5 Fascist. The maximum amount of elections to enact a policy is 3 due to the chaos mechanic. Either 2 elections fail and the 3\textsuperscript{rd} passes, or all three elections fail - the complexity is identical. There are 10 different ways for an election to fail (5 choose 3). There are 10 ways to choose a proposed government (5 choose 2). Hence, we have $(10 \times 10)^{3 \times 10} = 10^{60}$. This calculation does not take the hidden roles into effect, or any changes to the number of players due to presidential powers.\par

The size of the hidden state may also be of interest - this includes the policy deck and the hidden roles. In the 5 player version of the game, the 3 secret roles can be distributed to 5 players in 20 ways. Hitler can be any of the 5 players, and the other Fascist can be any of the remaining 4. The policy deck has 17 cards – 11 Fascist and 6 Liberal. The number of distinct decks can be calculated as $\frac{17!}{11!6!} = 12376$. Therefore, the total number of possible hidden states is $20 \times 12376 = 247520 \approx 10^5$

\section{Source Code}\label{app:source}

All source code for this project lives in a GitHub repository at \href{https://github.com/jackdreinhardt/secret-hitler-bots}{github.com/jackdreinhardt/secret-hitler-bots}. Python is the primary language used throughout.

\newpage

\section{SO-ISMCTS}\label{app:algorithms}
\footnotetext[1]{While the selection shown here is based on UCB, other bandit algorithms could be used instead.}

\begin{algorithm}[hp]
\DontPrintSemicolon
\SetKwProg{function}{function}{}{}
\SetKwFunction{soismcts}{SO-ISMCTS}\SetKwFunction{select}{Select}
\SetKwFunction{expand}{Expand}\SetKwFunction{simulate}{Simulate}
\SetKwFunction{backpropagate}{Backpropagate}
\function{\soismcts{$[s_0]^{\sim1}$, $n$}}{
    create single node tree with root $v_0 = [s_0]^{\sim1}$\;
    \For{$n$ iterations}{
        choose $d_0 \in [s_0]^{\sim1}$ uniformly at random\;
        $(v,d) \leftarrow$ \select{$v_0$, $d_0$}\;
        \If{$u(v,d) \neq \emptyset$}{
           $(v,d) \leftarrow$ \expand{$v$, $d$}\;
        }
        $r \leftarrow$ \simulate{$d$}\;
        \backpropagate{$r$, $v$}\;
    }
    \KwRet $a(c)$ where $c \in \argmax_{c \in c(v_0)} n(c)$\;
}
\function{\select{$v$, $d$}}{
    \While{$d$ is non-terminal and $u(v,d) = \emptyset$}{
    select\footnotemark[1] $c \in \argmax_{c \in c(v,d)} \left(\frac{r(c)_{\rho(d)}}{n(c)} + k\sqrt{\frac{\log n'(c)}{n(c)}}\right)$\;
    $v \leftarrow c$; $d \leftarrow f(d,a(c))$\;
    }
    \KwRet $(v, d)$
}
\function{\expand{$v$, $d$}}{
    choose $a$ from $u(v,d)$ uniformly at random\;
    add a child $w$ to $v$ with $a(w) = a$\;
    $v \leftarrow w$; $d \leftarrow f(d, a)$\;
    \KwRet $(v, d)$
}
\function{\simulate{$d$}}{
    \While{$d$ is non-terminal}{
        choose $a$ from $\Alpha(d)$ uniformly at random\;
        $d \leftarrow f(d, a)$\;
    }
    \KwRet $\mu(d)$
}
\function{\backpropagate{$r$, $v_l$}}{
    \ForEach{node $v$ from $v_l$ to $v_0$}{
        increment $n(v)$ by 1; $r(v) \leftarrow r(v) + r$\;
        let $d_v$ be the determinization when $v$ was visited\;
        \ForEach{sibling $w$ of $v$ compatible with $d_v$, including $v$ itself}{
            increment $n'(w)$ by 1\;
        }
    }
}

\caption[]{The SO-ISMCTS Algorithm taken from \cite{cowling_information_2012}. The following notation is used:
\begin{itemize}
    \setlength\multicolsep{0pt}
    \makeatletter%
    \if@twocolumn\else
        \begin{multicols}{2}
    \fi
    \item $c(v) =$ children of node $v$
    \item $a(v) =$ incoming action at node $v$
    \item $n(v) =$ visit count for node $v$
    \item $n'(v) =$ availability count for node $v$
    \item $r(v) =$ total reward for node $v$
    \makeatletter%
    \if@twocolumn\else
        \end{multicols}
    \fi
    \item $c(v, d) = \{u \in c(v):a(u) \in \Alpha(d)\}$, children of $v$ compatible with determinization $d$
    \item $u(v, d) = \{a \in \Alpha(d):\nexists c \in c(v, d)\;with\; a(c) = a\}$, the actions from $d$ for which $v$ does not have children in the current tree. Note that $c(v, d)$ and $u(v, d)$ are defined only for $v$ and $d$ such that $d$ is a determinization of (i.e., a state contained in) the information set to which $v$ corresponds.
\end{itemize}
}
\label{alg:soismcts}
\end{algorithm}

\end{document}